\documentclass[sigconf, nonacm=true]{acmart}

\usepackage{hyperref}
\usepackage{listings}
\usepackage{subcaption}

\AtBeginDocument{%
  \providecommand\BibTeX{{%
    \normalfont B\kern-0.5em{\scshape i\kern-0.25em b}\kern-0.8em\TeX}}}

\hypersetup{
    colorlinks=true,
    linkcolor=blue,
    filecolor=magenta,      
    urlcolor=cyan,
    pdftitle={Overleaf Example},
    pdfpagemode=FullScreen,
    }

\begin{document}

\title{MROS: A framework for robot self-adaptation}

\author{Gustavo Rezende Silva}
\authornote{Authors are with the Department of Cognitive Robotics, Delft University of Technology, Delft, Netherlands; Emails: g.rezendesilva@tudelft.nl, m.a.garzonOviedo@tudelft.nl, m.ramirezmontero-1@student.tudelft.nl, c.h.corbato@tudelft.nl}

\author{Darko Bozhinoski}
\authornote{Authors is with IRIDIA, University Libre du Bruxelles, Bruxelles, Belgium; Email: darko.bozhinoski@ulb.be}

\author{Mario Garzon Oviedo}
\authornotemark[1]

\author{Mariano Ramírez Montero}
\authornotemark[1]

% \affiliation{} %hack to split into 2 columns

\author{Nadia Hammoudeh Garcia}
\authornote{Authors are with Fraunhofer Institute for Manufacturing Engineering and Automation IPA, Stuttgart - Germany; Emails: nadia.hammoudeh.garcia@ipa.fraunhofer.de, harshavardhan.deshpande@ipa.fraunhofer.de}

\author{Harshavardhan Deshpande}
\authornotemark[3]

\author{Andrzej Wasowski}
\authornote{Author is with Computer Science Department IT University Conpenhagen, Denmark; Email: wasowski@itu.dk}

\author{Carlos Hernandez Corbato}
\authornotemark[1]

%% This command allows
%% the author to define a more concise list
%% of authors' names for this purpose.
\renewcommand{\shortauthors}{Gustavo Rezende, et al.}

\begin{abstract}
  Self-adaptation can be used in robotics to increase system robustness and reliability. This work describes the Metacontrol method for self-adaptation in robotics. Particularly, it details how the MROS (Metacontrol for ROS Systems) framework implements and packages Metacontrol, and it demonstrate how MROS can be applied in a navigation scenario where a mobile robot navigates in a factory floor. 
  %It also demonstrates how to reuse and apply MROS with a practical example. 
  Video: \url{https://www.youtube.com/watch?v=ISe9aMskJuE}
\end{abstract}


\maketitle

\section{Introduction}

Autonomous robots are usually programmed by integrating and configuring individual robot capabilities in a governing static control architecture. However, static control architectures fall short when addressing context variability in open-ended, dynamic  environments, where internal errors also compromise the quality and autonomy of mission execution. The demand for higher autonomy in robotic systems requires that robots can decide and switch between available alternative configurations, acting as a self-adaptive system. The addition of self-adaptation capabilities to robotic systems could be performed by including the adaptation logic into the robot's application logic. However, this leads to the problems: (1) the adaptation logic being tangled up with application logic, making it difficult to individually change any of the separate logic; (2) the lack of reuse of the adaptation reasoning, requiring it to be reprogrammed for each application. Thus, it is advantageous to use methods and tools that enable the separation of concerns and promote reuse. This paper demonstrate how Metacontrol \cite{corbato2013model} can be applied with the MROS \cite{bozhinoski2022mros} tooling to enable self-adaptation in robotic systems.

Metacontrol \cite{corbato2013model} is a framework that incorporates systems with the capability to self-adapt to maintain their functionalities at an expected performance. Metacontrol has the design goals of being reusable and extensible. This is achieved with the design principles: (1) separating the adaptation and application reasoning; (2) exploiting at runtime the engineering knowledge of how the system is designed to reason how and when the system needs to adapt, i.e., by being model-based.

MROS \cite{bozhinoski2022mros} is a tooling that implements the Metacontrol framework for ROS-based \cite{quigley2009ros} systems. This paper demonstrates how MROS\footnote{This paper address the ROS1 version of MROS, which can be found at: \url{https://github.com/meta-control/mc_mros_reasoner/tree/mros1-master}} lowers the barrier for robot developers to add self-adaptation to a robot control architecture by leveraging Metacontrol's design principles to promote reuse and extensibility.
In this paper, the novel contributions are:
\begin{itemize}
    \item A methodology for robotic developers to effectively use the MROS tooling to add self-adaptation capabilities in a robotic system.
    \item A case study that demonstrates the value of the MROS tooling and the proposed methodology.
\end{itemize}

\section{Metacontrol}\label{sec:metacontrol}
Metacontrol \cite{corbato2013model} is a reference architecture that provides systems with the capability to self-adapt to maintain their functionalities at an expected performance, despite external disturbances, faults, and unexpected behavior. For this, it uses engineering knowledge of how the system is designed, in the form of a runtime model, to reason when and how the system needs to adapt. Metacontrol is further explored in the works \cite{10.1007/978-3-319-22879-2_58, hernandez2018self, hernandez2019meta, aguado2021functional, bozhinoski2022mros, passler2022formal}.

\begin{figure}[h]
    \begin{minipage}[b]{0.6\linewidth}\centering
        \centering
        \includegraphics[width=0.9\linewidth]{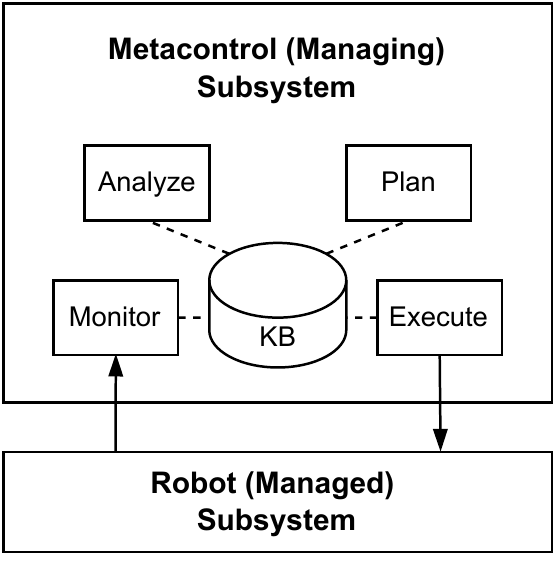}
        \caption{Metacontrol}
        \label{fig:metacontrol_pattern}
    \end{minipage}
    \begin{minipage}[b]{0.35\linewidth}\centering
        \centering
        \includegraphics[width=0.5\linewidth]{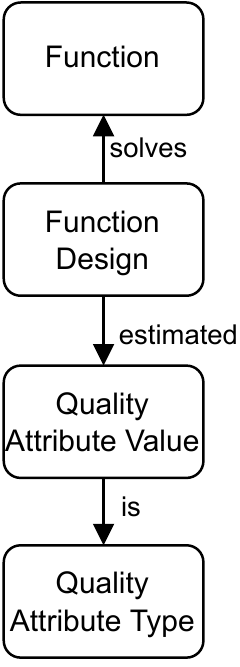}
        \caption{TOMASys}
        \label{fig:tomasys}
    \end{minipage}
\end{figure}

Metacontrol follows the standard model in self-adaptive systems (Figure \ref{fig:metacontrol_pattern}) that separates the application into a Managed system, which consists of the robotic domain specific components, and a Managing subsystem, which is a Metacontroller component that closes a feedback loop with the Robot subsystem to monitor and adapt it when necessary.

The Metacontroller operates following the MAPE-K loop \cite{kephart03computer}, which consists of a Monitor, Analyze, Plan and Execute steps, that are based on a Knowledge Base (KB). The Monitor step is responsible for measuring relevant quality attributes for the functionalities of the system. The Analyze step is responsible for deciding whether the Managed system needs to reconfigure. When it is necessary to perform reconfiguration, the Plan step is responsible for selecting a new configuration for the Managed system that satisfies its requirements. The Execute step is responsible for reconfiguring the Managed system.

The KB is the main difference of Metacontrol with other self-adaptation frameworks. It consists of a runtime model based on the TOMASys metamodel. TOMASys contains concepts to describe the functional and physical architecture of a system, and its variants, both at design time and at runtime. For the robotic developer using MROS, only TOMASys design time concepts are needed to create the Metacontrol KB, while the runtime elements are used by an automatic reasoner in the Metacontroller. A more detailed description of TOMASys can be found in \cite{corbato2013model, aguado2021functional}.

The \textit{Function} element represents an abstract functionality of the system, such as navigating from point A to B. A \textit{Function Design} is an engineering design solution that solves a specific \textit{Function} with an expected performance. \textit{Quality Attribute Type} and \textit{Quality Attribute Value} are used to capture the systems engineering meaning of QAs. A \textit{Quality Attribute Type} represents a characteristic of the system that shall be observed, such as energy. And a \textit{Quality Attribute Value} represents an amount of a \textit{Quality Attribute Type}, e.g., 1 Joule. 

\section{MROS framework}

MROS is a ROS-based implementation of Metacontrol that enables architectural self-adaptation in ROS-based robotic systems. MROS monitors the state of the ROS system and updates its KB according to it, then it uses ontological reasoning to Analyze and Plan the required adaptations, which are then executed by reconfiguring the ROS node graph. This section first describes how the MROS library implements the MAPE-K loop and the Metacontrol KB, and then it presents the methodology that robot developers can follow to use it to implement self-adaptation in ROS 1 applications\footnote{There is already a beta version of MROS supporting ROS 2, but it requires a different method}.

\subsection{MROS library}
\begin{figure*}
    \centering
    \includegraphics[width=0.65\linewidth]{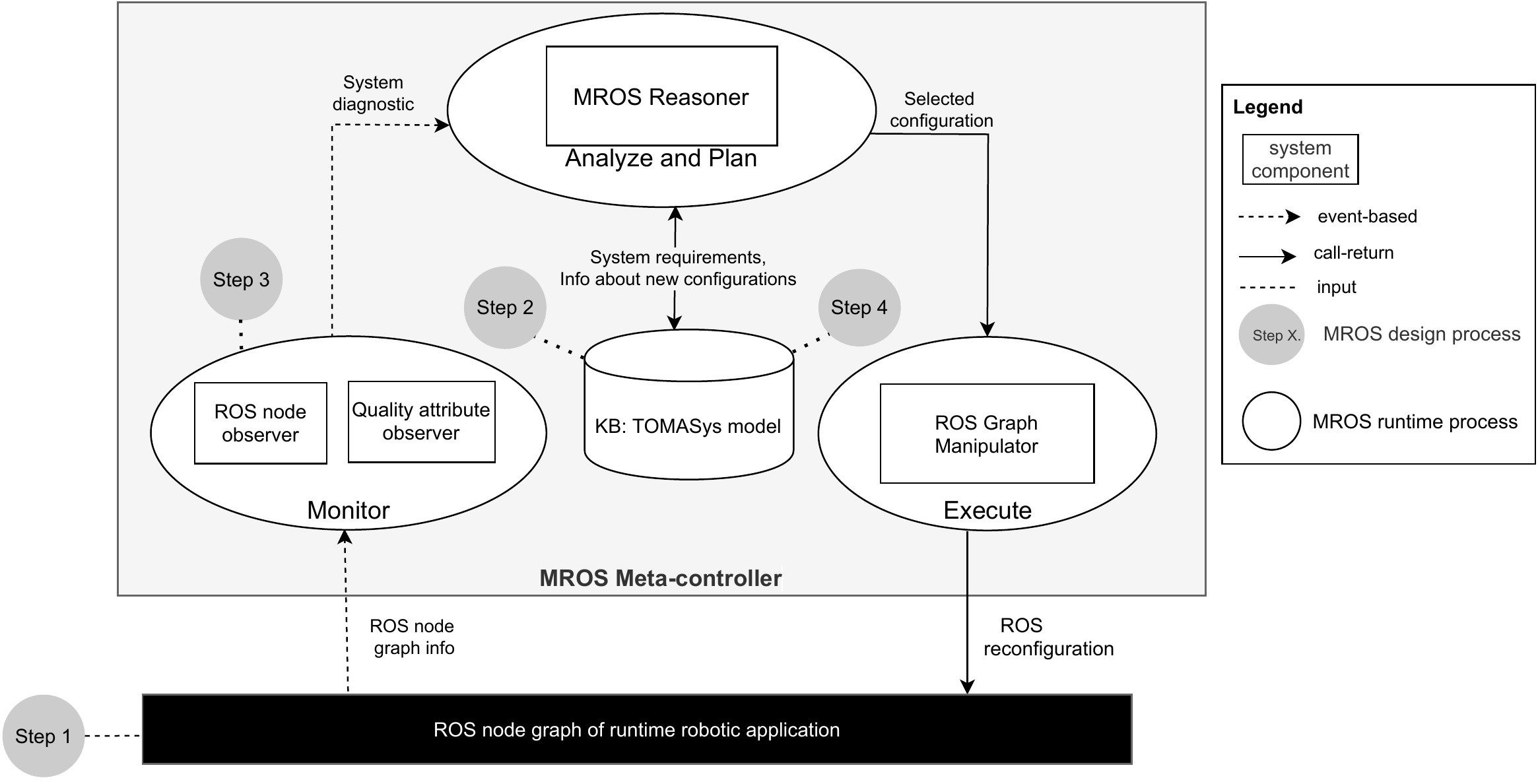}
    \caption{MROS framework}
    \label{fig:mape_loop}
\end{figure*}

MROS follows the MAPE-K model for runtime adaptation, as shown in in Figure \ref{fig:mape_loop}.

\textbf{Monitor}: This component is realized with the ROS node \href{https://github.com/rosin-project/rosgraph_monitor/tree/observers}{\emph{rosgraph\_monitor}}, it provides system observers templates. The observer nodes operate at a fixed frequency and publish the monitored values to the Metacontrol reasoner using the standard ROS diagnostic mechanism (i.e., a specific channel and data structure in ROS systems). This means they easily integrate into an existing ROS system and might use or provide services already needed in the system, regardless of the presence of a Metacontroller.

\textbf{Knowledge Base:} The Knowledge Base consists of \textbf{(1)} the TOMASys metamodel that allows modeling the functional architecture of autonomous systems; \textbf{(2)} A TOMASys model of the Managed subsystem that defines its unique \textit{Functions}, \textit{Function Designs} and its expected \textit{Quality Attributes}, in other words, all possible configurations for the Managed subsystem. The KB is implemented as an ontology using the Ontology Web Language (OWL) in combination with the Semantic Web Rule Language (SWRL).

\textbf{Analyze and Plan:} These components are realized with the ROS node \href{https://github.com/tud-cor/mc_mros_reasoner}{\emph{mros1\_reasoner}}, it integrates the OWL ontology, i.e. the KB, with ROS and it reasons over it to decide when and how to adapt the Managed subsystem.  It consists of a ROS node implemented with Python that makes use of the library OwlReady2 to bridge the ontology with Python and ROS. And the reasoning is performed with the off-the-shelf ontological reasoner Pellet.

\textbf{Execute:} This step is realized with the ROS node \\ \href{https://github.com/tud-cor/mc_rosgraph_manipulator}{\emph{mc\_rosgraph\_manipulator}}. It provides a ROS node that is responsible for killing and starting new ROS nodes and changing the necessary ROS parameters to fulfill the desired configuration.

\subsection{MROS methodology}\label{sec:methodology}

\begin{figure}[h]
    \centering
     \includegraphics[width=.8\linewidth]{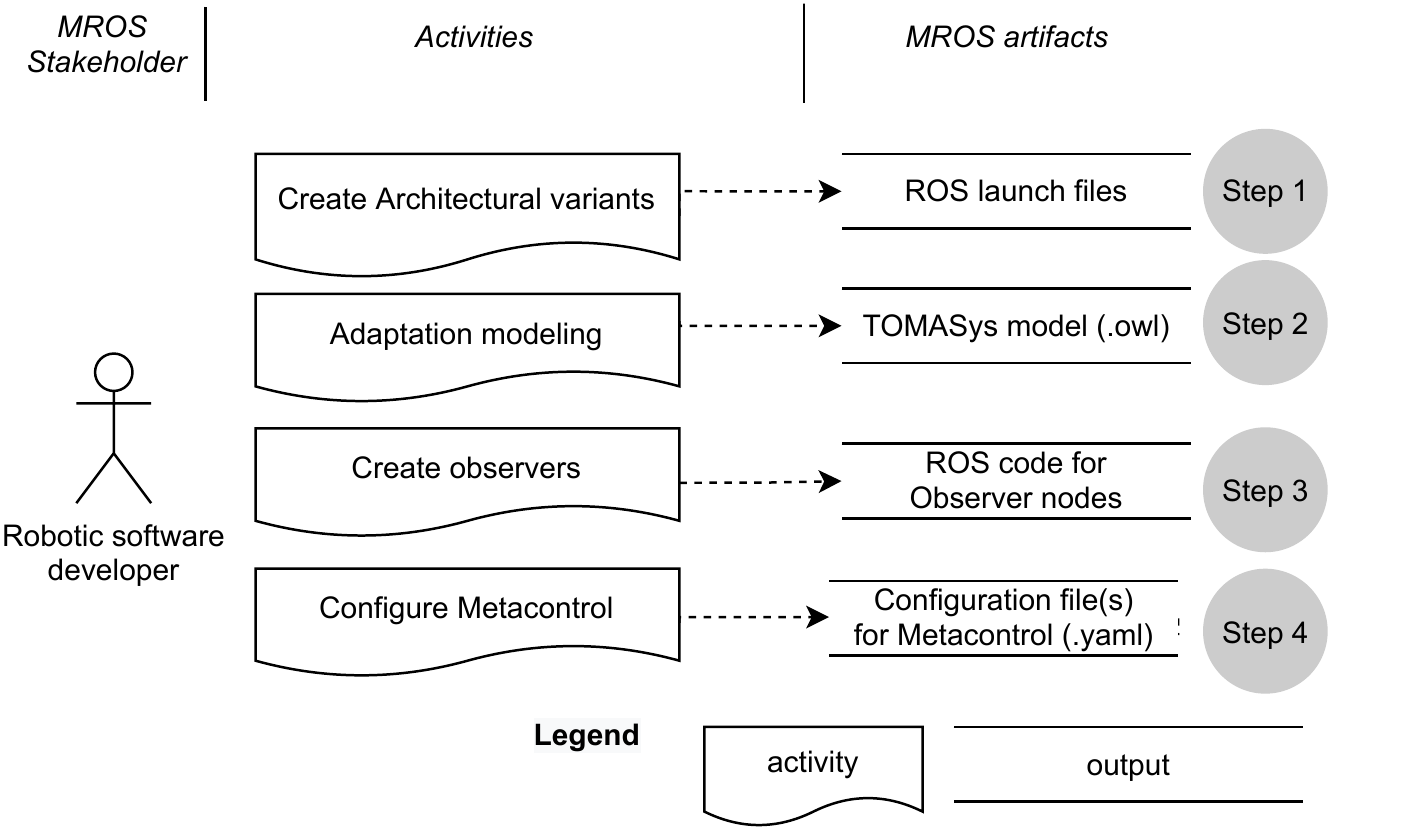}
    \caption{MROS Design time activities}
    \label{fig:design_time}
\end{figure}

To use MROS to add self-adaptation to a ROS system, the activities in Figure~\ref{fig:design_time} may be followed. 

\textbf{Step 1:} Define the possible adaptations, i.e. set of architectural variants relevant for the robot to be able to perform a mission. For each architectural variant, the developer creates a ROS launch file for the automatic deployment of the corresponding configuration of ROS nodes. 

\textbf{Step 2:} Create the KB by modeling the ROS system architectural variants with TOMASys. MROS provides an implementation of TOMASys in OWL, the developer needs to create the individuals for the application's  \textit{Functions}, \textit{Function Designs} and its expected \textit{Quality Attribute Values} and \textit{Quality Attribute Types}.

\textbf{Step 3:} Create observers using the templates provided in the ROS package \href{https://github.com/rosin-project/rosgraph_monitor/tree/observers}{\emph{rosgraph\_monitor}} to monitor the status of the active ROS nodes and to measure the relevant \textit{Quality Attributes}.
% , which are non-functional requirements of the system.

\textbf{Step 4:} Configure the \emph{mros1\_reasoner} and \emph{mc\_rosgraph\_manipulator} nodes  by linking the system architectural variants (step 1) with TOMASys \textit{Function Designs} (step 2), and connecting them to application-specific reconfiguration actions (e.g. to store required node states). This is done simply editing a template Metacontroller configuration file in YAML. 

\section{Case study}
This section describes how MROS can be set up for a robot navigating on a factory floor following the MROS methodology described in Section \ref{sec:methodology}. This case study comes for the previous work of Bozhinoski et al. \cite{bozhinoski2022mros}. The experiment code can be found at \url{https://github.com/rosin-project/metacontrol_sim}. 

\paragraph{Case study description}

The case study consists of a Clearpath Ridgeback mobile robot that navigates in a factory floor. The robot is equipped with two laser sensors, one IMU, and an odometry system.  The navigation system is realized with the ROS1 navigation stack. 
As the robot navigates, unexpected obstacles may appear in it's path and it may get closer or further to objects, which causes its safety quality attribute level to change.  When safety levels are high (lower risk of collision), the robot can navigate with higher speed and acceleration. When safety levels are low (higher risk of collision), it needs to use lower speeds. In addition, throughout the mission, the battery level diminishes and with this the robot must navigate with lower speed and acceleration to save energy. MROS is used to adapt at runtime the navigation parameters, such as maximum speed and acceleration, to satisfy safety and energy constraints. 

%Subsection MROS methodology?
\subsection{Application to MROS to the case study}

\textbf{Step 1:} Identify the architectural variants corresponding to different configurations (parameter values) of the main node of the ROS1 navigation stack, and create its corresponding launch files. In total, 27 different function designs were defined. A snippet of the ROS launch file of one \textit{Function Design} is shown in listing \ref{list:fd1}. The parameters that have been specifically defined for this \textit{Function Design} are: \textit{max\_vel\_x}, \textit{max\_vel\_y}, \textit{acc\_lim\_x}, \textit{acc\_lim\_y}, \textit{qa\_safety}, and \textit{qa\_energy}. The QA values are specified as ROS parameters to enable, if necessary, to change them at runtime. The QA values defined in the launchfiles override the ones defined in the ontology.

\begin{minipage}{0.95\linewidth}
\begin{lstlisting}[label={list:fd1}, caption=Function design 1 (out of 27), basicstyle=\ttfamily\scriptsize]
<launch>
  <param name="qa_safety" value="0.7"/>
  <param name="qa_energy" value="0.33"/>
  <node pkg="move_base" type="move_base" name="move_base" 
    cwd="node" respawn="false" output="screen">
    <param name="TrajectoryPlannerROS/max_vel_x" 
      value="0.3" />
    <param name="TrajectoryPlannerROS/max_vel_y" 
      value="0.3" />
    <param name="TrajectoryPlannerROS/acc_lim_x" 
      value="3.6" />
    <param name="TrajectoryPlannerROS/acc_lim_y" 
      value="3.6" />
  </node>
</launch>
\end{lstlisting}
\end{minipage}

\textbf{Step 2:} Model the architectural variants with the MROS TOMASys ontology. To create the OWL file for the KB, the graphical tool Protegé can be used to simplify the process. The TOMASys metamodel is available in the package \href{https://github.com/tud-cor/mc_mdl_tomasys}{mc\_mdl\_tomasys}. It is only necessary to set up the application-specific ontology by creating individuals of the design time TOMASys classes (Figure \ref{fig:tomasys}). 

For this case study, the following individuals are created: 
\begin{itemize}
    \item A \textit{Function} individual for the navigation capability;
    \item \textit{Quality Attribute Type} individuals for both safety and energy quality attributes;
    \item \textit{Function Design} individuals for each variant that solves navigation, including object property individuals of \textit{Quality Attribute Value} with their \textit{Quality Attribute Type} and a data field with the expected QA value.
\end{itemize}

\textbf{Step 3:} Create observers with the templates provided in the ROS package  \href{https://github.com/rosin-project/rosgraph_monitor/tree/observers}{rosgraph\_monitor}. These templates are two Python classes called \textit{TopicObserver} and \textit{ServiceObserver} that implement the general functionalities needed to monitor ROS topics and services, respectively. For each specific quality attribute that needs to be monitored, it is necessary to implement a new class that inherits from one of them.

For the use case, two Observers are implemented: \textit{SafetyQualityObserver} and \textit{EnergyQualityObserver}. A snippet of the implementation of the latter can be seen in listing \ref{list:observer}. The class \textit{EnergyQualityObserver} inherits from \textit{TopicObserver}. In the initialization of the class, the topic to which the observer needs to subscribe and its message type are defined. In line 4, it is defined that it needs to subscribe to the topic \textit{/power\_load} to retrieve information about the battery, and that its message type is a float. For all observers, the method \textit{calculate\_attribute} must be overloaded, it is responsible for performing any necessary calculation with the data received via the topic defined in the initialization, it must return the final data as a key-value pair structured as a \textit{diagnostic\_msgs/DiagnosticArray} message. In the \textit{EnergyQualityObserver}, in line 9, the method calculates the battery level as a normalized value, and returns it, for example, as $\{energy, 0.9\}$. The output of the observers is published in the topic \textit{diagnostics}.

\begin{minipage}{0.95\linewidth}
\begin{lstlisting}[language=Python,label={list:observer}, caption=EnergyQualityObserver Observer snippet, basicstyle=\ttfamily\scriptsize]
class EnergyQualityObserver(TopicObserver):
  def __init__(self, name):
    #topic to observe and msg type
    topics = [("/power_load", Float32)]
    super(EnergyQualityObserver,
        self).__init__(name, 10, topics)

  def calculate_attr(self, msgs):
      status_msg = DiagnosticStatus()
      
      #normalized calculus for energy
      attr =(msgs[0].data - 0.2)/(5.0-0.2)
      print("normalized energy: {0}".format(str(attr)))
      
      status_msg = DiagnosticStatus()
      status_msg.level = DiagnosticStatus.OK
      status_msg.name = self._id
      status_msg.values.append(KeyValue("energy", str(attr)))
      status_msg.message = "QA status"
      return status_msg
\end{lstlisting}
\end{minipage}

\textbf{Step 4:} Configure Metacontrol through a yaml file. This is used to map each \textit{Function Design} defined in the ontology (step 2) to its respective launch file (step 1). Additionally, to indicate which ROS nodes are killed and spawned during the reconfiguration process, as well, what actions and goals must be saved to be reset when the reconfiguration is performed. A snippet of the configuration file of this use case can be seen in listing \ref{list:yaml}.

\begin{minipage}{0.95\linewidth}
\begin{lstlisting}[label={list:yaml}, caption=Metacontrol configuration file, basicstyle=\ttfamily\scriptsize]
reconfiguration_action_name: 'rosgraph_manipulator_action_server'
configurations:
  f1_v1_r1:
    command: roslaunch f1_v1_r1 f1_v1_r1.launch
  f1_v1_r2:
    command: roslaunch f1_v1_r2 f1_v1_r2.launch
  f1_v1_r3:
    command: roslaunch f1_v1_r3 f1_v1_r3.launch
    
kill_nodes: ['/move_base']
save_action: 'move_base'
goal_msg_type: move_base_msgs.msg.MoveBaseAction

\end{lstlisting}
\end{minipage}

\subsection{Results}

Bozhinoski et al. \cite{bozhinoski2022mros} show that by adding self-adaptation with MROS to this use case, the robot performance increases regarding the amount of time it violates its required safety and energy quality attributes, and the overall mission success. In average, safety violations decreases from 2.5\% to 0.96\%, energy violations from 2.98\% to 1.86\%, and mission success increases from 65.20\% to 78.50\%.

\section{Related Work}

Aldrich et al. leverages predictive data models to enable automated robot adaptation to changes in the environment at run-time \cite{aldrich2019model}. While the approach depicts the benefits of using models by capturing high-level artifacts, it makes it extremely challenging for a ROS developer to make use of them in robotic scenarios because: (1) it does not introduce models that can be reused for a different application; (2) it does not give insights on how to build similar models; (3) it does not provide infrastructure to leverage those models.

Cheng et al. propose a framework that uses GSN assurance case models to manage run-time adaptations for ROS systems \cite{cheng2020ac}. The framework integrates assurance information from GSN models to ROS specific information to guide runtime monitoring and adaptation. It uses custom-developed libraries specific to the approach, rather than standard libraries in ROS (such as ROS Diagnostics) raising the entry barrier for ROS developers to effectively use it.

\section{Discussion and future works}

This paper describes MROS - a tool that enables robots to perform self-adaptation at runtime based on ontological reasoning. MROS establishes generic self-adaptation mechanisms that drive self-adaptation through the MAPE-K reference feedback loop. This eases the process of designing self-adaptation for robots since it only requires users to define the proper observers, the Managed system ontological model conforming to TOMASys, a few configuration files, and the launch files for each architectural variant. Due to its reusability and extensibility, MROS has been used to handle different adaptation concerns in different robotic applications, such as reliable propulsion and motion control in underwater robots \cite{aguado2021functional}, contingency handling in mobile manipulators \cite{bozhinoski2022mros}, and enhanced safety and energy saving in the navigation of a mobile robot \cite{bozhinoski2021context}.

% \begin{acks}
\paragraph{Acknowledgments}
This work was supported by the European Union's Horizon 2020 Framework Programme through the MSCA network REMARO (Grant Agreement No 956200), the grant RobMoSys-ITP-MROS (Grant Agreement No. 732410) and the ROSIN project (Grant Agreement No. 732287).
% \end{acks}

\bibliographystyle{ACM-Reference-Format}
\bibliography{references}

\end{document}